\documentclass[3p]{elsarticle}

\usepackage{amssymb}
\usepackage{booktabs}       % professional-quality tables
\usepackage{subfig}
\usepackage{graphicx}
\usepackage{multirow}
\usepackage{multicol}
\usepackage{array}
\usepackage{amsmath}

\usepackage{import}
\usepackage{xifthen}
\usepackage{pdfpages}
\usepackage{transparent}
\usepackage{bm}
\usepackage{amssymb}

\usepackage{listings}

\newcommand{\ca}[1]{\bm{\mathcal{#1}}}

\journal{Neurocomputing}

\begin{document}

\begin{frontmatter}



\title{TedNet: A Pytorch Toolkit for Tensor Decomposition Networks}


% \author{A. Author}

% \address{Your institute, some address}

\author[mymainaddress]{Yu Pan}
\ead{iperryuu@gmail.com}

\author[mythirdaddress]{Maolin Wang}
\ead{morin.w98@gmail.com}

\author[mymainaddress,mysecondaryaddress]{Zenglin Xu\corref{mycorrespondingauthor}}
\cortext[mycorrespondingauthor]{Corresponding author}
\ead{zenglin@gmail.com}

\address[mymainaddress]{Harbin Institute of Technology Shenzhen, Shenzhen, China}
\address[mysecondaryaddress]{Pengcheng Lab, Shenzhen, China}
\address[mythirdaddress]{University of Electronic Science and Technology of China, Chengdu, China}

% \begin{abstract}
% %% Text of abstract 
% Ca. 100 words

% \end{abstract}

% \begin{keyword}
% %% keywords here, in the form: keyword \sep keyword
% keyword 1 \sep keyword 2 \sep keyword 3

% %% PACS codes here, in the form: \PACS code \sep code

% %% MSC codes here, in the form: \MSC code \sep code
% %% or \MSC[2008] code \sep code (2000 is the default)

% \end{keyword}

\begin{abstract}%   <- trailing '%' for backward compatibility of .sty file

Tensor Decomposition Networks (TDNs) prevail for their inherent compact architectures. To give more researchers a flexible way to exploit TDNs, we present a Pytorch toolkit named TedNet. TedNet implements 5 kinds of tensor decomposition(i.e., CANDECOMP/PARAFAC (CP), Block-Term Tucker (BTT), Tucker-2, Tensor Train  (TT) and Tensor Ring (TR) on traditional deep neural layers, the convolutional layer and the fully-connected layer. By utilizing the basic layers, it is simple to construct a variety of TDNs. TedNet is available at https://github.com/tnbar/tednet.
\end{abstract}

% \begin{keywords}
%   Tensor Decomposition Networks, Deep Neural Network, Tensor Optimization
% \end{keywords}
\begin{keyword}
Tensor Decomposition Networks \sep Deep Neural Networks \sep Tensor Networks \sep Network Compression
\end{keyword}

\end{frontmatter}

% \linenumbers


\section{Introduction}

Tensor Decomposition Networks (TDNs) are constructed by decomposing deep neural layers with tensor formats. For the reason that the original tensor of a layer can be recovered from tensor decomposition cores, TDNs are often regarded as a compression method for the corresponding networks. Compared with traditional networks like Convolution Neural Networks (CNNs) and Recurrent Neural Networks (RNNs), TDNs can be much smaller and occupy a little memory. For example, TT-LSTM~\cite{DBLP:conf/icml/YangKT17}, BTT-LSTM~\cite{DBLP:conf/cvpr/YeWLCZCX18,DBLP:journals/nn/YeLCYZX20}, TR-LSTM~\cite{DBLP:conf/aaai/PanXWYWBX19,li2021heuristic} are able to reduce  17,554, 17,414 and 34,192 times parameters with a higher accuracy than the original models. With light-weight architectures and good performance,  TDNs are promising to be used in kinds of source-restricted scenes including mobile equipment and microcomputers.  Due to these advantages, TDNs can often achieve comparably high accuracy with huge parameter reduction in a number of tasks, such as action recognition~\cite{DBLP:journals/pieee/PanagakisKCONAZ21,sun2020human}. TDNs have also been implemented in FPGA for fast inference with ultra memory reduction~\cite{zhang2021fpga} and multi-task learning to improve the representing ability~\cite{wang2020concatenated}. Under this background, we design TedNet package for providing convenience for researchers to explore on TDNs.

\begin{figure}[t]
\centering
\includegraphics[width=0.9\columnwidth]{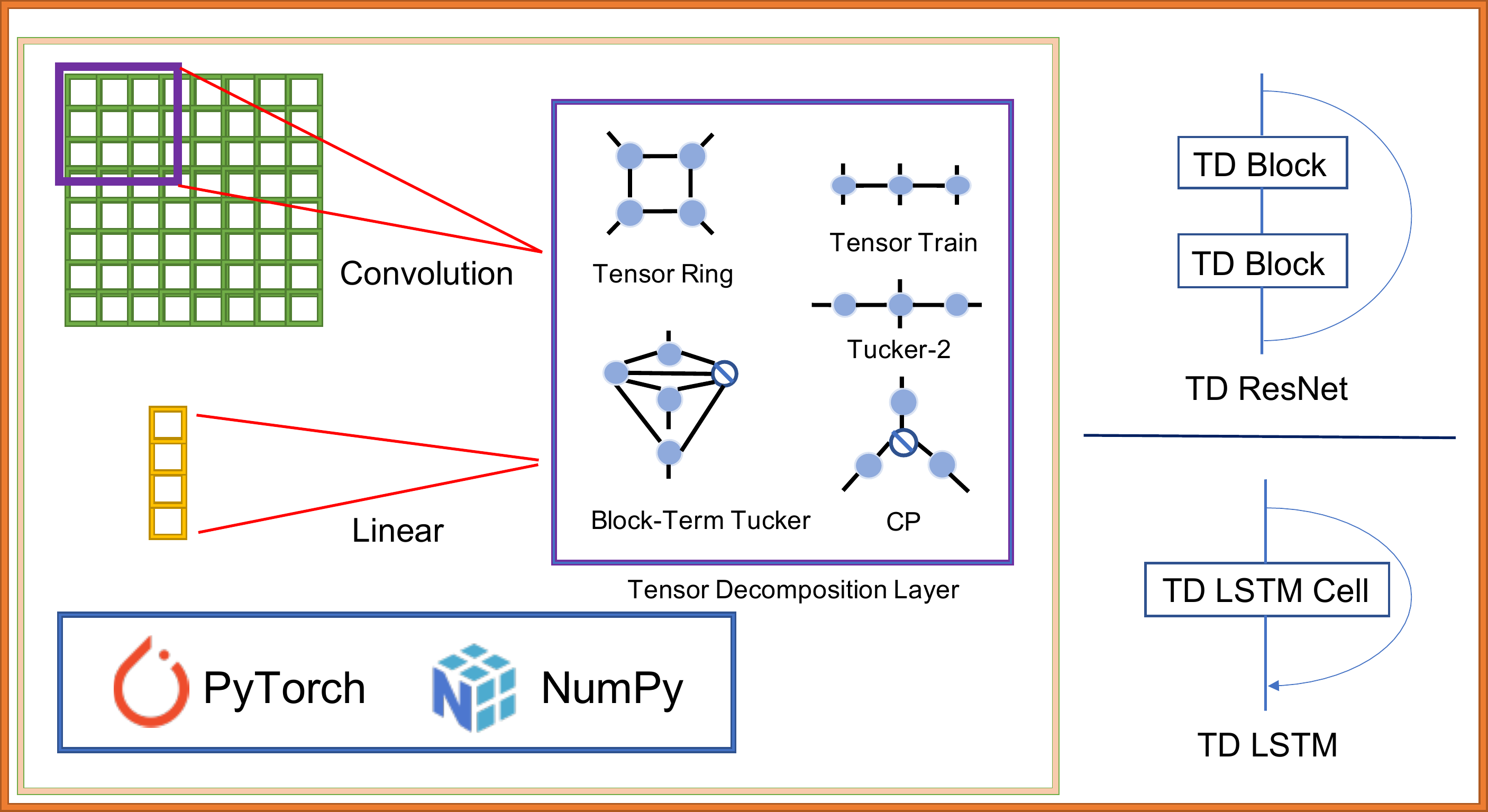}
\caption{The framework of TedNet.  TedNet is based on Pytorch and adopts NumPy to process numerical calculations. Tensor decomposition (TD) can be applied to convolutional layers or linear layers.  We implemented 5 variants of tensor decomposition methods, namely CP, Tucker, Tensor Ring, Tensor Train, and Block-term Tucker. Tensor decomposition can be fulfilled in convolution neural networks. An illustration of two tensorial classical neural blocks(i.e., ResNet and  LSTM) that are built on the Tensor Decomposition Layer is shown in the right of the figure.}
\label{fig:frame}
\end{figure}

There are several related packages, such as T3F~\cite{DBLP:journals/jmlr/NovikovIKFO20}, Tensorly~\cite{DBLP:journals/jmlr/KossaifiPAP19}, TensorD~\cite{DBLP:journals/ijon/HaoLYX18}, TensorNetwork~\cite{roberts2019tensornetwork}, tntorch~\cite{tntorch} , OSTD~\cite{DBLP:conf/iccvw/SobralJJBZ15} and TensorTools~\cite{williams2018unsupervised}.  OSTD is constructed for low-rank decomposition and implemented with MATLAB. TensorTools based on NumPy~\cite{DBLP:journals/cse/WaltCV11} implements CP decomposition only, while T3F is explicitly designed for Tensor Train Decomposition on Tensorflow~\cite{DBLP:conf/osdi/AbadiBCCDDDGIIK16}. Similarly based on Tensorflow, TensorD supports CP and Tucker decomposition. By contrast, TedNet implements five kinds of tensor decomposition with backend Pytorch~\cite{NEURIPS2019_9015}.  TensorNetwork is built on Tensorflow and incorporates abundant tensor calculation tools. Nevertheless, TensorNetwork serves for tensor decomposition algorithms rather than TDNs. Tensorly supports with a variety of backends including CuPy, Pytorch, Tensorflow and MXNet~\cite{chen2015mxnet}. Unfortunately, although Tensorly is powerful to process tensor algebra, tensor decomposition and tensor regressions, it still lacks support to Application Programming Interface (API) to build tensorial neural networks directly. Interestingly, Tensorly can assist to initialize TedNet network modules with its tensor decomposition operation. Compared with them, TedNet can 
set up a TDN layer quickly by calling API directly. In addition, we also provide three kinds of deep TDNs that are popular for researchers now. Due to the Dynamic Graph Mechanism of Pytorch, TedNet is also flexible to DEBUG for programmers.

% {\noindent \em Remainder omitted in this sample. See http://www.jmlr.org/papers/ for full paper.}

% Acknowledgements should go at the end, before appendices and references


\section{TedNet Details}

TedNet is designed with the goal of building TDNs by calling corresponding APIs, which can extremely simplify the process of constructing TDNs. As shown in Figure~\ref{fig:frame}, TedNet adopts Pytorch as the training framework because of its auto differential function and convenience to build DNN models.  In addition, TedNet also uses NumPy~\cite{DBLP:journals/cse/WaltCV11} to assist in tensor operations. The fundamental module of TedNet is \textbf{\_TNBase}, which is an abstracted class and inherits from \textbf{torch.nn.Module}. Thus, TedNet models can be amicably combined with other Pytorch models. As an abstracted class,  \textbf{\_TNBase} requires sub-classes to implement 4 functions~\footnote{https://github.com/tnbar/tednet/blob/main/tednet/tnn/tn\_module.py}.  On the right side of Figure~\ref{fig:frame}, we show two main deep architectures of TedNet, namely TD ResNet and TD LSTM, which are  probably the most frequently used backbone in convolutional neural networks and recurrent neural networks, respectively.

% \begin{table}[t]
% \centering
% \begin{tabular}{l|l} 
% \hline
% Function                                          & Description                                                       \\ 
% \hline\hline
%  set\_tn\_type     & Set the tensor decomposition type.                                \\ 
% \hline
%  set\_nodes                        & \begin{tabular}[c]{@{}l@{}}Generate tensor decomposition nodes, \\ then edit node information.\end{tabular}  \\ 
% \hline
%  set\_params\_info & Record information of Parameters.                                 \\ 
% \hline
%  tn\_contract                      & The function of contracting inputs and tensor nodes.              \\ 
% \hline
% %  recover                                           & Be used for rebuilding the original tensor.                       \\
% % \hline
% \end{tabular}
% \caption{Functions of \textbf{\_TNBase}.}
% \label{tbl:tnbase}
% \end{table}

Usually, DNNs are constructed with CNNs and Linears. The weight of a CNN is a 4-mode tensor $\ca{C} \in \mathbb{R}^{K\times K \times C_{in} \times C_{out}}$, where $K$ means the convolutional window, $C_{in}$ denotes the input channel and $C_{out}$ represents the counterpart output channel. And a Linear is a matrix $\mathbf{W} \in \mathbb{R}^{I \times O}$, where $I$ and $O$ are length of input and output feature respectively. Similar to DNNs, TDNs consist of TD-CNNs and TD-Linears(For simplification, TD- denotes the corresponding tensor decomposition model), whose weights $\ca{C}$ and $\textbf{W}$ are factorized with tensor decomposition.  Following this pattern, 
there are 5 frequently-used tensor decomposition (i.e. CP, Tucker-2, Block-Term Tucker, Tensor Train and Tensor Ring) in TedNet, which satisfies most of common situations. Notably,  TedNet is an open-source package which supports Tensor Ring Decomposition. Besides, based on TD-CNNs and TD-Linears, TedNet has built some tensor decomposition based Deep Neural Networks, e.g. TD-ResNets, TD-RNNs.

\begin{lstlisting}[language={Python}, label={ll:tr}, caption={A demo of applying Tensor Ring Decomposition for MNIST Classification.}]
# Import Tensor Ring Module for Calling
import tednet.tnn.tensor_ring as tr

# Import Necessary Pytorch Modules
import torch
import torch.nn as nn
from torch import Tensor


# A Simple MNIST Classifier based on Tensor Ring.
class TRClassifier(nn.Module):
    def __init__(self):
        super(TRClassifier, self).__init__()

        # Define a Tensor Ring Convolutional Layer
        self.tr_cnn = tr.TRConv2D([1], [4, 5], [6, 6, 6, 6], 3)
        # Define a Tensor Ring Fully-Connected Layer
        self.tr_fc = tr.TRLinear([20, 26, 26], [10], [6, 6, 6, 6])

    def forward(self, inputs: Tensor) -> Tensor:
        # Call TRConv2D to process inputs
        out = self.tr_cnn(inputs)
        out = torch.relu(out)
        out = out.view(inputs.size(0), -1)

        # Call TRLinear to classify the features
        out = self.tr_fc(out)
        return out
\end{lstlisting}

\section{Installation and Illustrative Examples}

There are two ways to install TedNet. For the sake that the source code of TedNet is submitted to GitHub, it is feasible to install from the downloaded code by command \textbf{python setup.py install}. Compared with aforementioned fussy way, another one, the recommended way is to install TedNet trough PyPI~\footnote{https://pypi.org/project/tednet} by command \textbf{pip install tednet}. After installation, all tensor decomposition models of TedNet can be used.

 A simple MNIST~\cite{lecunGradientbasedLearningApplied1998a} classifier based on tensor ring is shown in Listing~\ref{ll:tr}. The tensor ring module can be used by importing \textbf{tednet.tnn.tensor\_ring}. We utilize two fundamental tensor ring layers (i.e., TRConv2D, TRLinear) to build the sample classifier. In addition, it is very convenient to build a whole tensor ring network with only one line of code, e.g., TR-LeNet5~\footnote{https://tednet.readthedocs.io/en/latest/quick\_start.html}. The usage of other decomposition is the same and more details can be found in the Document~\footnote{https://tednet.readthedocs.io}.

% \begin{figure}[h]
% \centering
% \subfloat[Tensor Ring Models]{
%     % \caption{Interest region}
%     \includegraphics[width=0.48\columnwidth]{figures/tr-usage.png}
%     \label{fig:tr-usage}
% }
% \subfloat[Tensor Train Models]{
%     % \caption{Performance}
%     \includegraphics[width=0.48\columnwidth]{figures/tt-usage.png}
%     \label{fig:tt-usage}
% }
% \caption{Usage of TedNet.}
% \label{fig:usage}
% \end{figure}

% \lstinputlisting[
%     % style       =   Python,
%     caption     =   {\bf Tensor Ring Models},
%     label       =   {cd:tr}
% ]{codes/tr.py}

\section{Benchmark}

\begin{figure}[t]
\centering
\subfloat[Training Process in 150 epochs.]{
\includegraphics[width=0.45\columnwidth]{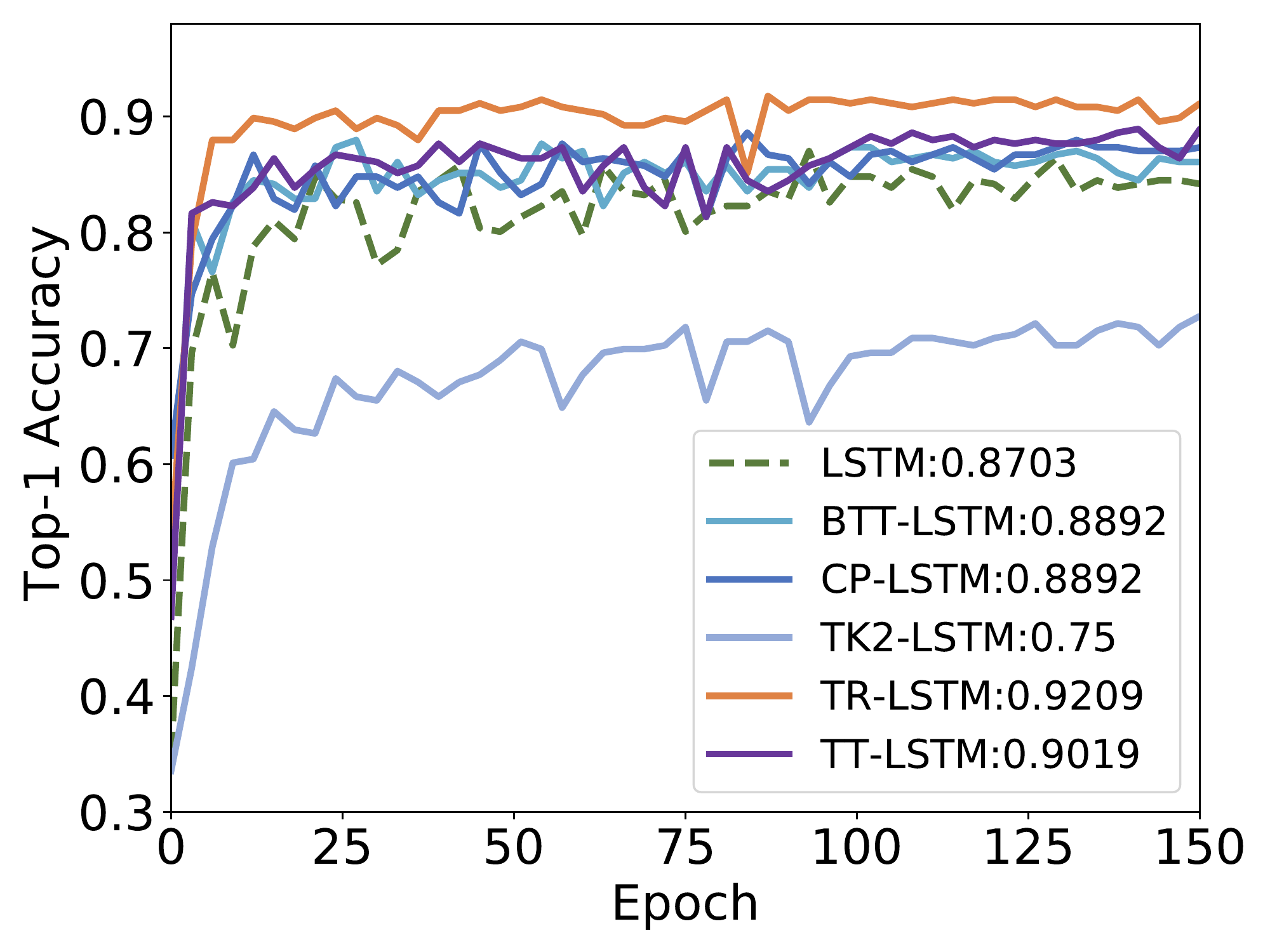}
}~~~~
\subfloat[Compression Ratio.]{
\includegraphics[width=0.45\columnwidth]{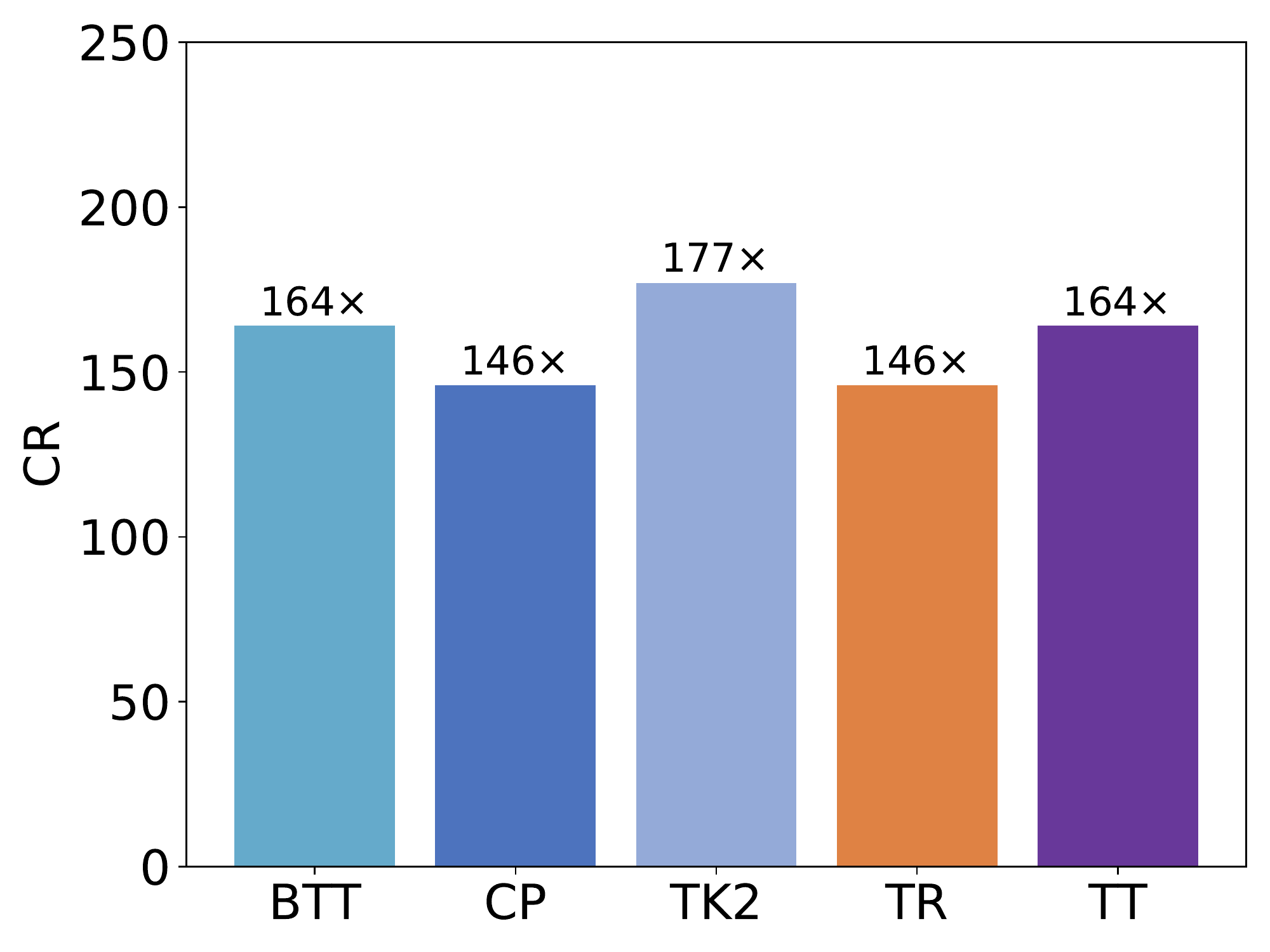}
}
\caption{Experiments on UCF11. CR is short for Compression Ratio.}
\label{fig:ucf11}
\end{figure}

Until now, TDNs are mostly applied in computer vision field. Thus, aiming to validate performance of TedNet, we consider to conduct experiments on two datasets:
\begin{itemize}
    \item The UCF11 Dataset contains  1,600 video clips of a resolution $320 \times 240$ and is divided into 11 action categories. Each category consists of 25 groups of videos, within more than 4 clips in one group. 
    \item The Cifar10/100 consists of 50,000 train images and 10,000 test images with size as $32 \times 32 \times 3$. CIFAR10 has 10 object classes and CIFAR100 has 100 categories.
\end{itemize}

 For the video classification task on UCF11, we adopt the same setting as described in literature \cite{DBLP:conf/aaai/PanXWYWBX19}, where  we extract feature of dimension 2048 from each frame of a video by Inception-V3~\cite{DBLP:conf/cvpr/SzegedyVISW16}. Then throw these features as step inputs into TD-LSTMs. Results are shown in Figure~\ref{fig:ucf11}. Almost every tensor decomposition model can achieve better accuracy except Tucker-2.

% \begin{table}[h]
% \centering
% \begin{tabular}{|l|l|l|l|l|} 
% \hline
% Model    & Rank & Params & CR & Accuracy  \\ 
% \hline
% LSTM     & - & 16777216 & 1$\times$   & 0.8861    \\ 
% \hline
% BTT-LSTM & 5  & 102450 & 164$\times$  & 0.8892    \\ 
% \hline
% CP-LSTM  & 400 & 115200 & 146$\times$ & 0.8892    \\ 
% \hline
% TK2-LSTM & 10 & 94880 & 177$\times$  & 0.75      \\ 
% \hline
% TR-LSTM  & 20 & 115200 & 146$\times$  & 0.9209    \\ 
% \hline
% TT-LSTM  & 10 & 102400 & 164$\times$  & 0.9019    \\
% \hline
% \end{tabular}
% \caption{Experiments on UCF11}
% \label{tbl:ucf11}
% \end{table}

% \begin{table}[h]
% \centering
% \begin{tabular}{l|c|c|c|c} 
% \hline
% Model    & Rank & Params & CR & Accuracy  \\ 
% \hline
% \hline
% LSTM     & - & 17M & 1$\times$0.8703    \\ 
% \hline
% BTT-LSTM & 5  & 0.10M & 164$\times$  & 0.8892    \\ 
% \hline
% CP-LSTM  & 400 & 0.12M & 146$\times$ & 0.8892    \\ 
% \hline
% TK2-LSTM & 10 & 0.09M & 177$\times$  & 0.75      \\ 
% \hline
% TR-LSTM  & 20 & 0.12M & 146$\times$  & \textbf{0.9209}    \\ 
% \hline
% TT-LSTM  & 10 & 0.10M & 164$\times$  & 0.9019    \\
% \hline
% \end{tabular}
% \caption{Experiments on UCF11. CR is short for Compression Ratio.}
% \label{tbl:ucf11}
% \end{table}

% As for Cifar10/100, we construct the image classification task on them. Through applying TD-ResNet-32, we can validate performance of all the tensor decomposition in TedNet. Results are demonstrated in Table~\ref{tbl:cifar}. Unlike the results of UCF11, tensor decomposition can lead to a negligible loss in contrast to the original accuracy.
 For the image classification task on Cifar10/100, we employ ResNet-32 as the backbone network.  We show the results of  corresponding TD-ResNet-32 implementations with various tensor decomposition in Table~\ref{tbl:cifar}.

 Note that the results shown in Table~\ref{tbl:cifar} and Figure~\ref{fig:ucf11} are obtained without fine tuning parameters, and are just used for verifying the correctness of these algorithms.  Thus the classification results does not mean the performance of these algorithms with the best parameter settings.

% \begin{table}[h]
% \centering
% \begin{tabular}{l|l|l|l|l|l|l|l} 
% \hline
% \multicolumn{2}{l|}{} & \multicolumn{3}{l|}{Cifar10} & \multicolumn{3}{l}{Cifar100}  \\ 
% \hline
% \hline
% Model         & Rank   & Params & CR & Acc.          & Params & CR & Acc.            \\ 
% \hline
% ResNet-32     & -      & 464154 &  1$\times$  & 0.9228         & 470004 &  1$\times$  & 0.6804           \\ 
% \hline
% BTT-ResNet-32 & 4      & 76410  & 6$\times$   & 0.8589         & 76788  &  6$\times$  & 0.5206           \\ 
% \hline
% CP-ResNet-32  & 10     & 26492  &  18$\times$  & 0.8802         & 26672  &  18$\times$  & 0.4445           \\ 
% \hline
% TK2-ResNet-32 & 10     & 54032  &  9$\times$  & 0.8915         & 55022  &  9$\times$  & 0.5398           \\ 
% \hline
% TR-ResNet-32  & 10     & 93082  &  5$\times$  & 0.9076         & 94172  & 5$\times$   & 0.653            \\ 
% \hline
% TT-ResNet-32  & 10     & 92852  &  5$\times$  & 0.9020         & 96542  &  5$\times$  & 0.6386           \\
% \hline
% \end{tabular}
% \caption{Experiments on Cifar.}
% \label{tbl:cifar}
% \end{table}

\begin{table}[t]
\centering
\scalebox{0.9}{
\begin{tabular}{l|c|c|c|c|c|c|c} 
\hline
\multicolumn{2}{l|}{} & \multicolumn{3}{c|}{Cifar10} & \multicolumn{3}{c}{Cifar100}  \\ 
\hline
\hline
Model         & Rank   & Params & CR & Accuracy          & Params & CR & Accuracy            \\ 
\hline
ResNet-32     & -      & 0.46M &  1$\times$  & 0.9228         & 0.47M &  1$\times$  & 0.6804           \\ 
\hline
BTT-ResNet-32 & 4      & 0.08M  & 6$\times$   & 0.8955         & 0.08M  &  6$\times$  & 0.5661           \\ 
\hline
CP-ResNet-32  & 10     & 0.03M  &  18$\times$  & 0.8802         & 0.03M  &  18$\times$  & 0.4445           \\ 
\hline
TK2-ResNet-32 & 10     & 0.05M  &  9$\times$  & 0.8915         & 0.06M  &  9$\times$  & 0.5398           \\ 
\hline
TR-ResNet-32  & 10     & 0.09M  &  5$\times$  & 0.9076         & 0.09M  & 5$\times$   & 0.653            \\ 
\hline
TT-ResNet-32  & 10     & 0.09M  &  5$\times$  & 0.9020         & 0.10M  &  5$\times$  & 0.6386           \\
\hline
\end{tabular}
}
\caption{Experiments on Cifar10/100. Params denotes the number of parameters.  CR means compression rate.  The number of block-terms is set to 5 in BTT-ResNet-32.}
\label{tbl:cifar}
\end{table}

\section{Conclusion}
In this paper, we present a package named TedNet that is specially designed for TDNs.  TedNet is completely open-source and distributed under the MIT license. Compared with other related python packages, TedNet contains the most kinds of tensor decomposition.

\section*{Acknowledgements}
% \label{}

This paper was partially supported by the National Key Research and Development Program of China (No. 2018AAA0100204), and a key  program of fundamental research from Shenzhen Science and Technology Innovation Commission (No. JCYJ20200109113403826).



%% References: At least 5 are required 
\bibliographystyle{elsarticle-num} 
\bibliography{mybibfile}





% \end{thebibliography}

\clearpage

\section*{Required Metadata}

% \section*{Current executable software version}

% Ancillary data table required for sub version of the executable software: (x.1, x.2 etc.) kindly replace examples in right column with the correct information about your executables, and leave the left column as it is.

% \begin{table}[!h]
% \begin{tabular}{|l|p{6.5cm}|p{6.5cm}|}
% \hline
% \textbf{Nr.} & \textbf{(executable) Software metadata description} & \textbf{Please fill in this column} \\
% \hline
% S1 & Current software version & 0.1.3 \\
% \hline
% S2 & Permanent link to executables of this version  &  $https://github.com/tnbar/tednet/$ $releases/tag/0.1.3$ \\
% \hline
% S3 & Legal Software License & MIT License \\
% \hline
% S4 & Computing platform/Operating System & Linux, OS X, Microsoft Windows \\
% \hline
% S5 & Installation requirements \& dependencies & Pytorch\\
% \hline
% S6 & If available, link to user manual - if formally published include a reference to the publication in the reference list & $https://tednet.readthedocs.io/$ $en/latest/index.html$ \\
% \hline
% S7 & Support email for questions & iperryuu@gmail.com\\
% \hline
% \end{tabular}
% \caption{Software metadata (optional)}
% \label{} 
% \end{table}

\section*{Current code version}
\label{}

Ancillary data table required for subversion of the codebase. Kindly replace examples in right column with the correct information about your current code, and leave the left column as it is.

\begin{table}[!h]
\begin{tabular}{|l|p{6.5cm}|p{8.2cm}|}
\hline
\textbf{Nr.} & \textbf{Code metadata description} & \textbf{Please fill in this column} \\
\hline
C1 & Current code version & 0.1.3 \\
\hline
C2 & Permanent link to code/repository used of this code version & $https://github.com/tnbar/tednet/releases/tag/0.1.3$ \\
\hline
C3 & Legal Code License   & MIT License \\
\hline
C4 & Code versioning system used & git \\
\hline
C5 & Software code languages, tools, and services used & Python, Pytorch \\
\hline
C6 & Compilation requirements, operating environments \& dependencies & Python3.X, NumPy\\
\hline
C7 & If available Link to developer documentation/manual & $https://tednet.readthedocs.io/en/latest/index.html$ \\
\hline
C8 & Support email for questions & iperryuu@gmail.com\\
\hline
\end{tabular}
\caption{Code metadata (mandatory)}
% \label{} 
\end{table}

\end{document}